\documentclass[10pt,twocolumn,letterpaper]{article}
\usepackage[utf8]{inputenc}
\usepackage{fancyhdr}
\usepackage{lmodern}			
\usepackage[T1]{fontenc}		
\usepackage{indentfirst}		
\usepackage{color}			
\usepackage{graphicx}			
\usepackage{microtype} 			
\usepackage[portuguese]{babel}
\usepackage{amsbsy}			
\usepackage{amsfonts}			
\usepackage{amsmath}			
\usepackage{amssymb}			
\usepackage{amstext}			
\usepackage{model}
\usepackage{times}
\usepackage{epsfig}
\usepackage{graphicx}
\usepackage{amsmath}
\usepackage{amssymb}
\usepackage{color}
\usepackage{comment}
\usepackage{colortbl}
\usepackage{multirow}
\usepackage{array}
\usepackage[dvipsnames,table]{xcolor}
\usepackage{pdflscape}
\usepackage{comment}
\usepackage[final]{pdfpages}
\usepackage{afterpage}
\usepackage{tikz}
\usepackage{smartdiagram}
\usetikzlibrary{matrix, positioning}
\usepackage{float}

\definecolor{bblue}{rgb}{0.0, 0.3, 0.9}

\definecolor{ggray}{rgb}{0.8, 0.8, 0.8}
\usepackage[pagebackref=true,breaklinks=true,letterpaper=true,colorlinks,bookmarks=false]{hyperref}

\newcommand{\ABACODP}[1]{%
\thicklines
\begin{picture}(8,0)
    \ifcase#1{   
       \put(0,0)    {\line(1,0){4}}
       \multiput(5,0)(2,0){2}{\oval(2,4)}}
    \or{         
       \put(2,0)    {\line(1,0){4}}
       \multiput(1,0)(6,0){2}{\oval(2,4)}}
    \fi
\end{picture}
    } 

\newcommand{\ABACODG}[1]{%
\thicklines
\begin{picture}(14,0)
    \ifcase#1{   
       \multiput(1,0)(2,0){5}{\oval(2,4)}}
       \put(10,0)   {\line(1,0){4}}
    \or{         
       \multiput(1,0)(2,0){4}{\oval(2,4)}}
       \put(8,0)   {\line(1,0){4}}
       \put(13,0)   {\oval(2,4)}
    \or{         
       \multiput(1,0)(2,0){3}{\oval(2,4)}
       \put(6,0)   {\line(1,0){4}}
       \multiput(11,0)(2,0){2}{\oval(2,4)}}
    \or{         
       \multiput(1,0)(2,0){2}{\oval(2,4)}
       \put(4,0)   {\line(1,0){4}}
       \multiput(9,0)(2,0){3}{\oval(2,4)}}
    \or{         
       \put(1,0)  {\oval(2,4)}}
       \put(2,0)   {\line(1,0){4}}
       \multiput(7,0)(2,0){4}{\oval(2,4)}
    \fi
\end{picture}
    } 
       
\newcommand{\ABACOD}[1]{%
    \ifnum#1>9
       \errmessage{#1: Argumento invalido para ABACO}
    \fi
    \ifnum#1<0
       \errmessage{#1: Argumento invalido para ABACO}
    \fi
\begin{picture}(24,0)
    \ifnum#1<5
       \put(16,0) {\ABACODP{0}}
    \else   
       \put(16,0) {\ABACODP{1}}
    \fi
    \ifnum#1<5
       \put(0,0)  {\ABACODG{#1}}
    \else
       \ifcase#1\or \or \or \or
          \or  \put(0,0)  {\ABACODG{0}}
          \or  \put(0,0)  {\ABACODG{1}}
          \or  \put(0,0)  {\ABACODG{2}}
          \or  \put(0,0)  {\ABACODG{3}}
          \or  \put(0,0)  {\ABACODG{4}}
       \fi
    \fi   
\end{picture}
    } 
    
\cvprfinalcopy 

\ifcvprfinal\fi

\begin{document}

\title{Predição de propriedades de ligas de aço}
\author{Ciro Javier Diaz Penedo\thanks{Do Instituto de Matemática, Estatística e Computação Científica da Universidade de Campinas (Unicamp). \textbf{Contato}: \tt\small{ra153868@ime.unicamp.br  e  ra153866@ime.unicamp.br}} \ - ra: 153868 \\
Lucas Leonardo Silveira Costa$^{*}$ \ - ra: 153866}

\maketitle
\begin{abstract}
Apresentamos um estudo de possíveis preditores baseados em quatro modelos de aprendizado de máquina supervisionado para a predição de quatro propriedades mecânicas dos principais aços utilizados industrialmente. Os resultados foram obtidos a partir de um banco de dados experimentais disponível na literatura os quais foram utilizados como input para treinar e avaliar os modelos. 
\end{abstract}

\section{Introdução}

Na área de engenharia de materiais não se conhece um modelo matemático para calcular algumas propriedades das ligas de aço, desse modo, a maneira mais usual para encontrar as mesmas é via experimentos em laboratório, os quais demandam especialistas, equipamentos, gastos financeiros e tempo. Uma opção para contornar esses gastos é prever o valor dessas propriedades utilizando algoritmos de aprendizado de máquina \cite{2}.

Na composição da liga de aço cada composto químico fornece uma propriedade diferente. As ligas metálicas existentes no mercado visam atender um grupo específico destas propriedades.

Atualmente existem \textit{handbooks} \cite{4} com informações sobre as propriedades físicas de ligas de aço. Esses bancos de dados fornecem apenas as características de ligas que já foram estudadas em laboratório previamente. 

Neste trabalho usamos vários modelos de aprendizado de máquina para analisar quatro propriedades de ligas de aço, dureza (\textit{hardness}), limite de resistência (\textit{tensile strength}), limite de escoamento (\textit{yield strength}) e ductibilidade (\textit{elongation in} 50mm$\%$). Estas propriedades dependem da composição química da liga e do tipo de processamento utilizado em sua fabricação: Laminação a quente, Laminação a frio, Recozimento, Normalização e Tempera em água ou óleo. A descrição das componentes das amostras são apresentadas na Tabela \ref{TabComp}.

\begin{table}
\centering
\scalebox{0.8}{ 
\begin{tabular}{|c|c|l|r|r|r|r|r|} 
\hline
\rowcolor{cyan}
$\#$ & sigla & Atributo &  $\%$ min & $\%$ média & $\%$ max \\  
\hline
\hline
1 & (Fe) & Ferro & 70 & 93,094  & 100,0  \\
\hline
\rowcolor{ggray}
2 & (C) & Carbono  & 0,08  & 0,355  & 1,2  \\
\hline
3 & (Mn) &	Manganês  & 0,25  & 0,971  & 2,0  \\
\hline
\rowcolor{ggray}
4 & (P) & Fósforo  &  0  & 0,028  & 0,2 \\
\hline
5 & (S) & Enxofre & 0  &  0,050  &  1,0 \\
\hline
\rowcolor{ggray}
6 & (Si) &	Silício  & 0  &  0,240  &  3,0\\
\hline
7 & (Ni) &	Níquel  & 0	 &  3,350  &  26,0\\
\hline
\rowcolor{ggray}
8 & (Cr) & Cromo  & 0 & 1,802  & 37,0 \\
\hline
9 & (Mo) &	Molibdênio   &  0   & 0,111 & 4,0\\
\hline
\rowcolor{ggray}
10 & $P$  & Preparo (1-5) & 1 & 2.375 &  5 \\
\hline
\end{tabular}}
\vspace{2mm}

\scalebox{0.85}{ 
\begin{tabular}{|l|l||l|l|r|r|} 
\hline
\rowcolor{cyan}
$P$ & Preparo & $P$ & Preparo \\  
\hline
1 & Laminação a quente & 2 & Laminação a frio  \\
\hline
\rowcolor{ggray}
3 & Recozimento & 4 & Normalização \\
\hline
5 & Têmpera em água ou óleo & & \\
\hline
\end{tabular}}
\caption{Composição química e tipo de preparo da liga metálica} \label{TabComp}
\label{TamComp}
\end{table}

 Os valores de cada elemento químico da Tabela \ref{TabComp} junto com a medida categórica (tipo de preparo) na parte de baixo serão as \textit{features} dos exemplo do problema e os \textit{targets} serão as quatro propriedades físicas-mecânica das ligas: \textit{hardness}, \textit{tensile strength}, \textit{yield strength }e \textit{elongation}. Todas elas foram retiradas do handbook \cite{4}. São um total de $207$ dados sendo as concentrações de cada elemento químico se encontram numa faixa de valores, e desse modo, foi possível realizar um \textit{data augmentation} ficando com $7952$ amostras.

\section{Metodología}

   Para obter nosso preditor vamos testar quatro modelos clássicos de aprendizado supervisionado, estes são \textit{Linear Regression} (LR), \textit{Redes Neurais} (NN), \textit{Regression Suport Vector Machines} (SVR) e \textit{Decision Trees} (DT). 

\subsection{\textit{Data augmentation}}\label{sub21}
  Para treinar os modelos de aprendizado precisamos de mais dados. A peculiaridade do problema tratado faz possível aumentar essa quantidade. Sendo que algumas propriedades são obtidas para faixas de concentração de elementos químicos decidimos usar vários valores dentro de cada faixa. Desta forma tomaremos 3 valores em cada faixa dividindo esta em 4 pontos equidistantes e ficamos com os 2 mais próximos do centro para treino e validação. Como conjunto de teste usaremos apenas o valor central de cada faixa. (Espera-se que as faixas sejam melhor representadas por valores mais próximos do centro do que da fronteira). 

   Fazendo o que foi citado anteriormente passamos a ter um conjunto de $7952$ amostras para treino e validação, e um conjunto  de $207$ amostras para teste, o qual foi usado apenas apara apresentar os resultados. 

\subsection{\textit{Métricas Avaliação}}\label{Bonf}

Neste trabalho vamos usar o coeficiente de determinação \textit{R-square} ($R^{2}$) para medir a qualidade da predição, sendo que $0 \leq R^{2} \leq 1$ e quanto mais próximo de $1$ melhor o resultado. 
  $$ \mbox{SQ}_{tot} = \sum_{i=1}^{n} (T_{i} - \bar{T}_{i})^{2}\hspace*{5mm} , \hspace*{5mm} \mbox{SQ}_{res} = \sum_{i=1}^{n} (T_{i} - y_{i})^{2}\hspace*{5mm} , $$
$$ \mbox{\textbf{R}}^{2} = 1 - \dfrac{\mbox{SQ}_{res}}{\mbox{SQ}_{tot}} \hspace*{5mm}, \hspace*{5mm} \mbox{\textbf{EQM}} = \dfrac{\mbox{SQ}_{res}}{n}.$$

  Para avaliar a capacidade de generalização de cada modelo a partir do conjunto de dados usaremos o método \textit{K-fold} de validação cruzada. Vamos usar $K = 10$ mantendo os mesmos $10$ conjuntos de treino e validação para cada modelo.

   Para comparar estatisticamente vários modelos (como vai ser o caso dos obtidos via SVR e NN) usamos o teste de \textit{Friedman} para determinar se existe ou não diferença estatística no conjunto total e o teste de \textit{Bomferroni} para fazer comparação \textit{one vs. one} (OVO) e determinar que existe diferencia estatística entre cada dois deles, esta existe quando a diferença dos \textit{rankings} é maior do que um valor critico,

\begin{equation*}
|r_i - r_j | > t_{\infty,k,\alpha}\cdot\sqrt[]{2\frac{EQM}{n}}
\end{equation*}

No caso usamos a função \textit{multicompare} do \textit{MATALB} que entrega uma matriz com as comparações OVO para as colunas de uma matriz de estatísticas. Usamos $\alpha=0.05$ no cálculo do valor crítico.

\section{Experimentos e Discussões}

Fazendo o \textit{data augmentation} como explicado em \ref{sub21}, treinamos cada modelo com a estratégia \textit{K-Folder} de validação cruzada evitando \textit{overfitting}. O resultado final é apresentado avaliando cada um dos $10$ preditores obtidos no conjunto de teste e promediando os resultados.

\subsection{\textit{Linear Regression}} 

  Vamos fazer ajuste dos dados a uma combinação linear de polinômios de ordem até $3$. Usando o método de quadrados mínimos e calculando os coeficientes via solução do sistema de equações normais conseguimos ajustar o grau dos polinômios e o parâmetro de regularização $\lambda$ para escolher nosso modelo de regressão Linear. Usamos a regra do cotovelo para decidir não aumentar a mais a complexidade do modelo no ponto onde erro no treino e a validação começam a se distanciar.

\begin{figure}[H]
\centering
\includegraphics[scale=0.5]{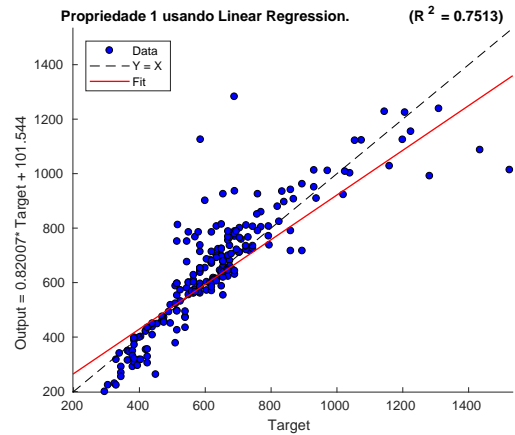}
\caption{Previsões \textit{vs targets} para a propriedade 1 (\textit{Linear Regression}).} \label{FigR1}
\end{figure}

\begin{figure}[H]
\centering
\includegraphics[scale=0.5]{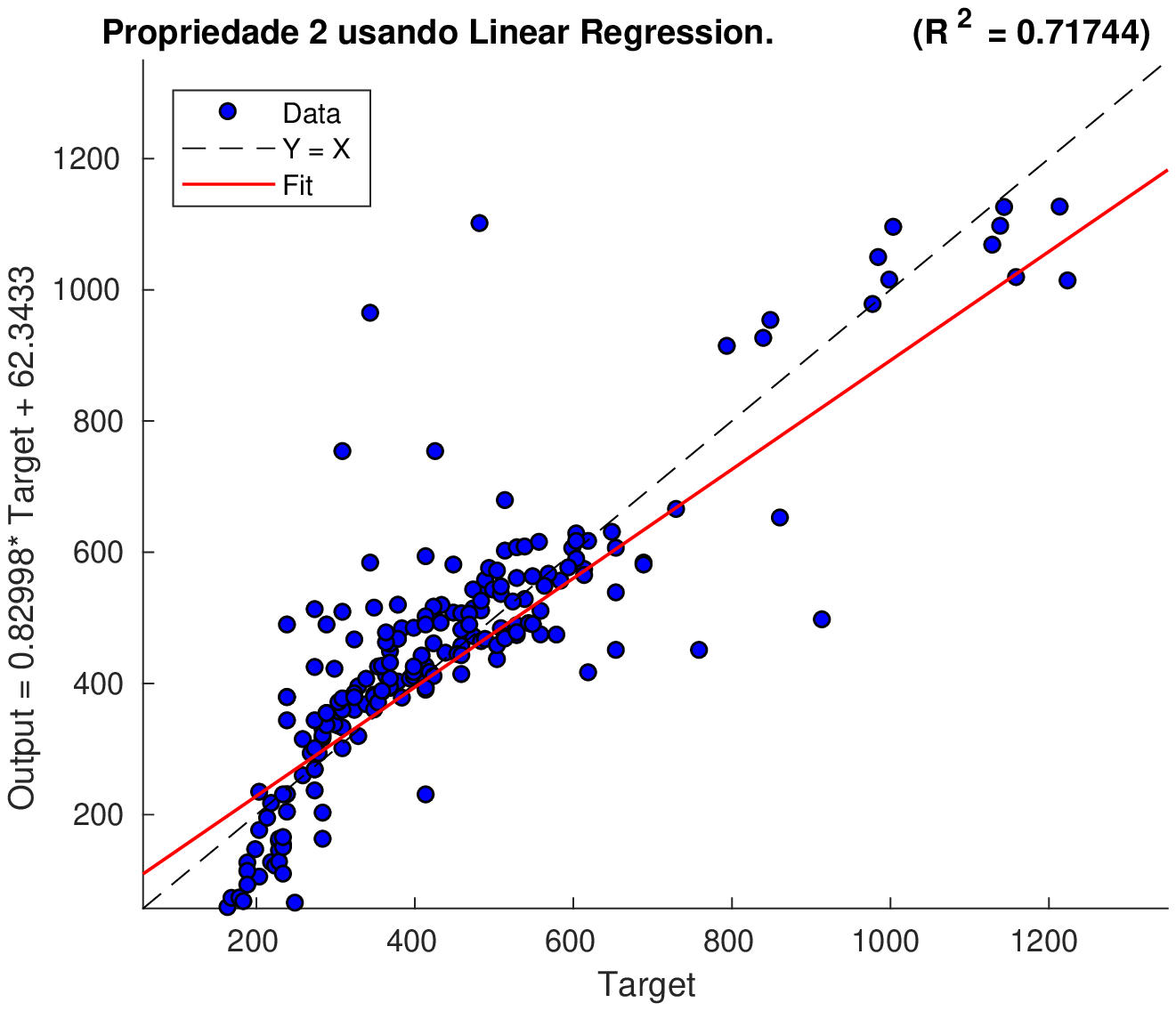} 
\caption{Previsões \textit{vs targets} para a propriedade 2 (\textit{Linear Regression}).}\label{FigR2}
\end{figure}

\begin{figure}[H]
\centering
\includegraphics[scale=0.5]{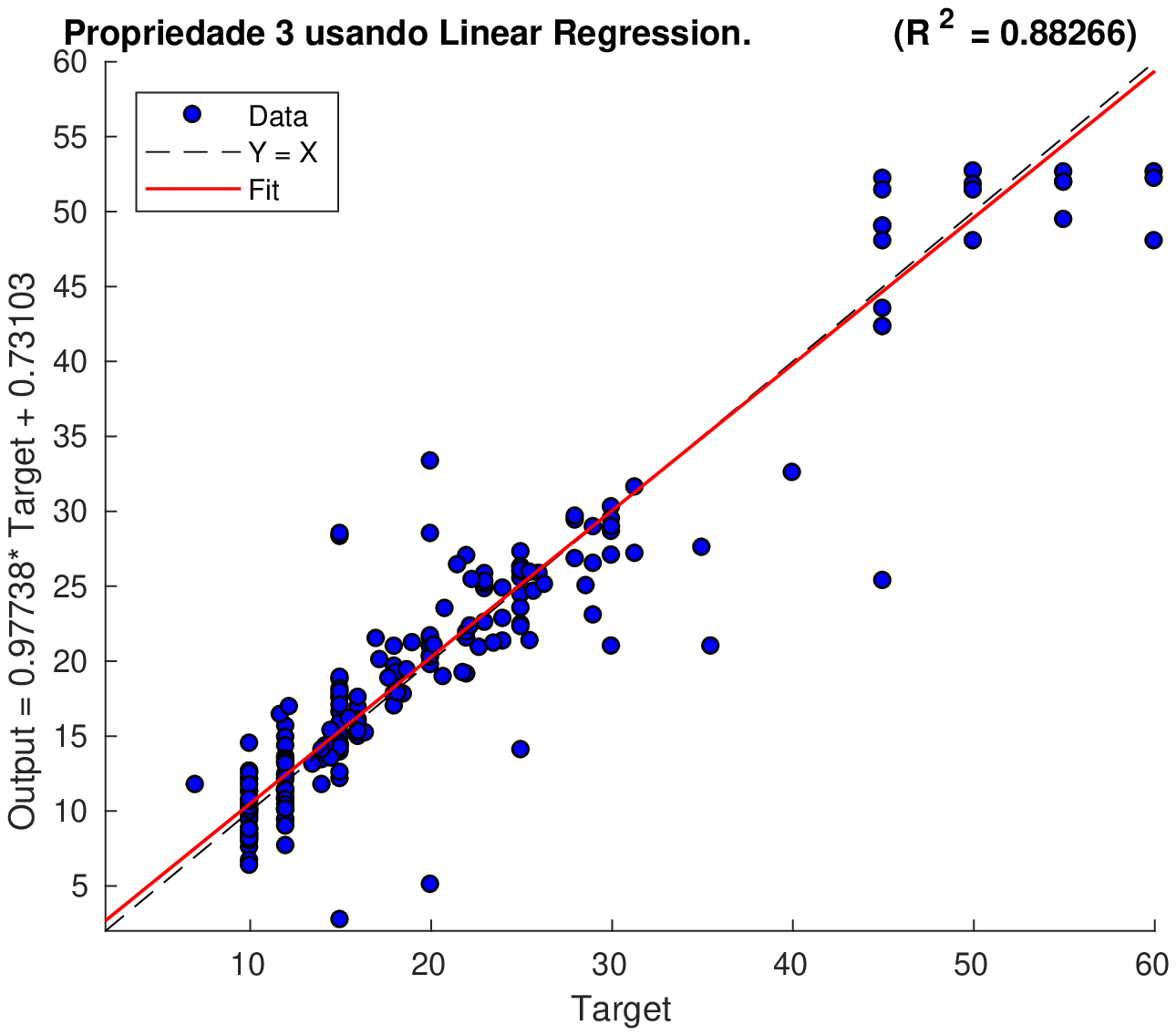}
\caption{Previsões \textit{vs targets} para a propriedade 3 (\textit{Linear Regression}).} \label{FigR3}
\end{figure}

\begin{figure}[H]
\centering
\includegraphics[scale=0.5]{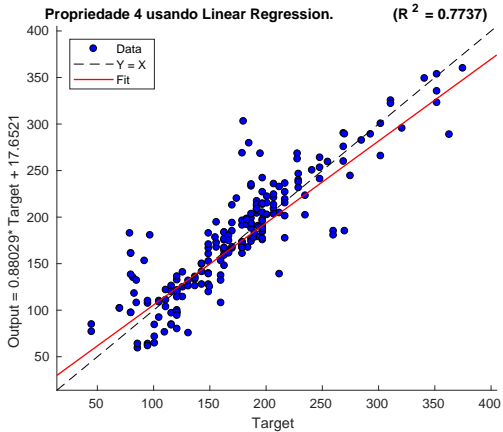}
\caption{Previsões \textit{vs targets} para a propriedade 4 (\textit{Linear Regression}).} \label{FigR4}
\end{figure}

 Os gráficos das Figuras \ref{FigR1}, \ref{FigR2}, \ref{FigR3} e  \ref{FigR4} apresentam a dispersão entre as previsões e os \textit{targets} para o conjunto de teste, para cada propriedade respectivamente. Quanto mais próximos os pontos se encontram da diagonal ($Y=X$) melhor foi o resultado, pois o $output = y_{i} \approx T_{i} = target$, quando mais distantes pior foi a previsão.

\subsection{\textit{Neural Network}} 

Para o treinamento das Redes Neurais usamos a função de \textit{MATLAB} \textit{fitnet}. A arquitetura da rede consiste de uma camada escondida na qual variamos a quantidade de neurônios $(1-10)$. Treinamos a rede variando o tipo de função de ativação usada no treino $(1-12)$: \textit{trainlm, trainbr, trainscg, trainbfg, traingdx, traingd, trainrp, traincgb, traincgf, traincgp, trainoss e  traingdm}. Para mais informações \cite{6}. 

\begin{figure}[H]
\centering
\includegraphics[scale=0.5]{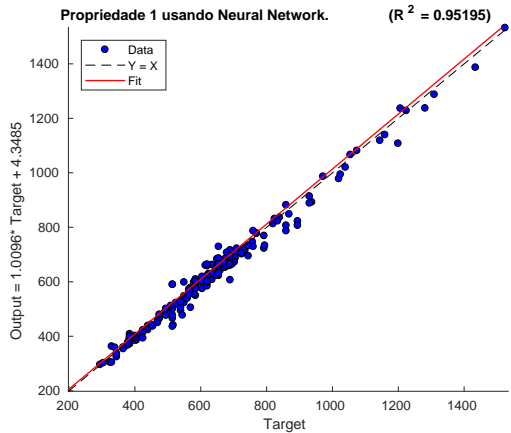}
\caption{Previsões \textit{vs targets} para a propriedade 1 (\textit{Neural Network}).} \label{FigNN1}
\end{figure}

\begin{figure}[H]
\centering
\includegraphics[scale=0.5]{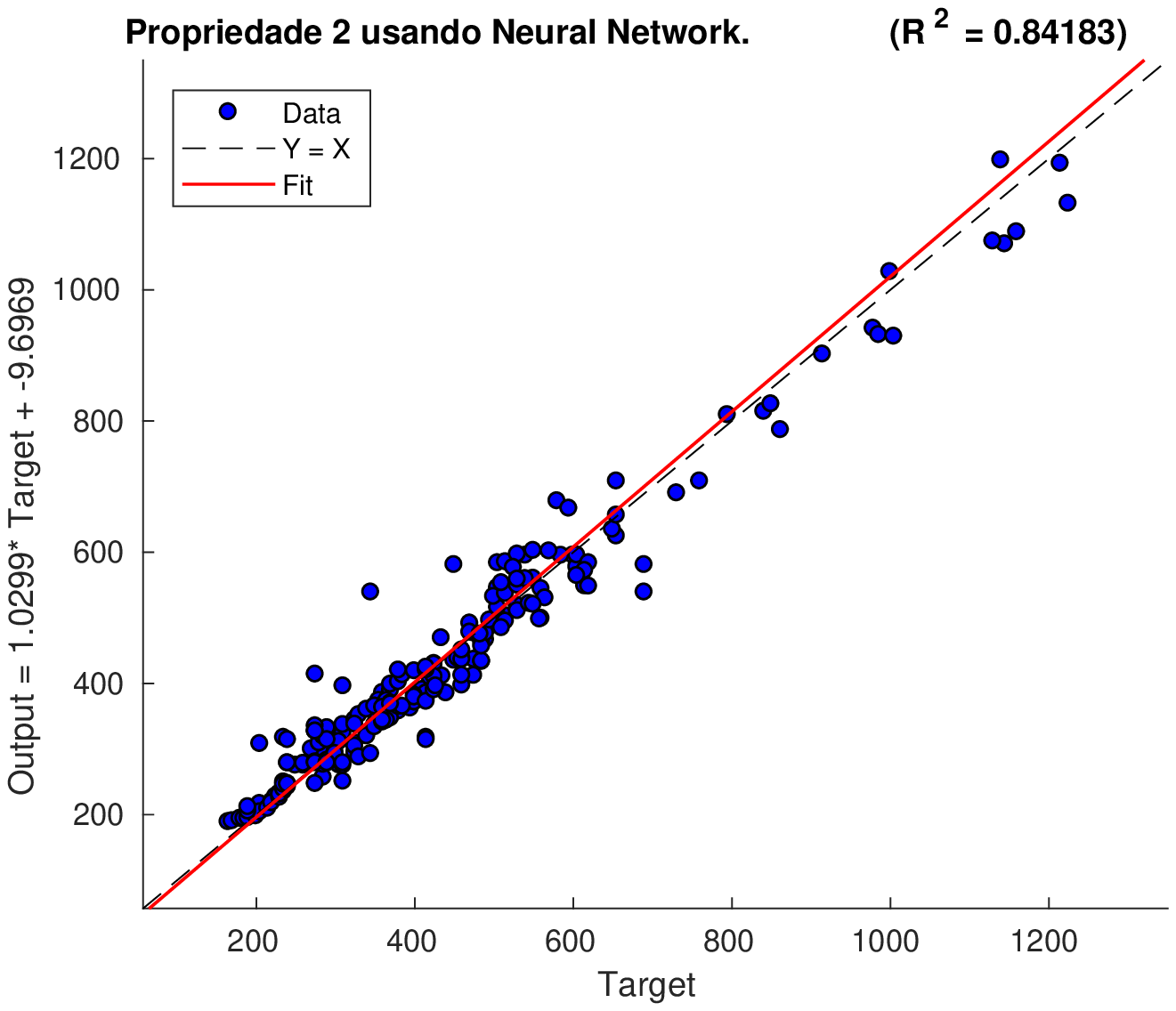} 
\caption{Previsões \textit{vs targets} para a propriedade 2 (\textit{Neural Networ}k).} \label{FigNN2}
\end{figure}

\begin{figure}
\centering
\includegraphics[scale=0.5]{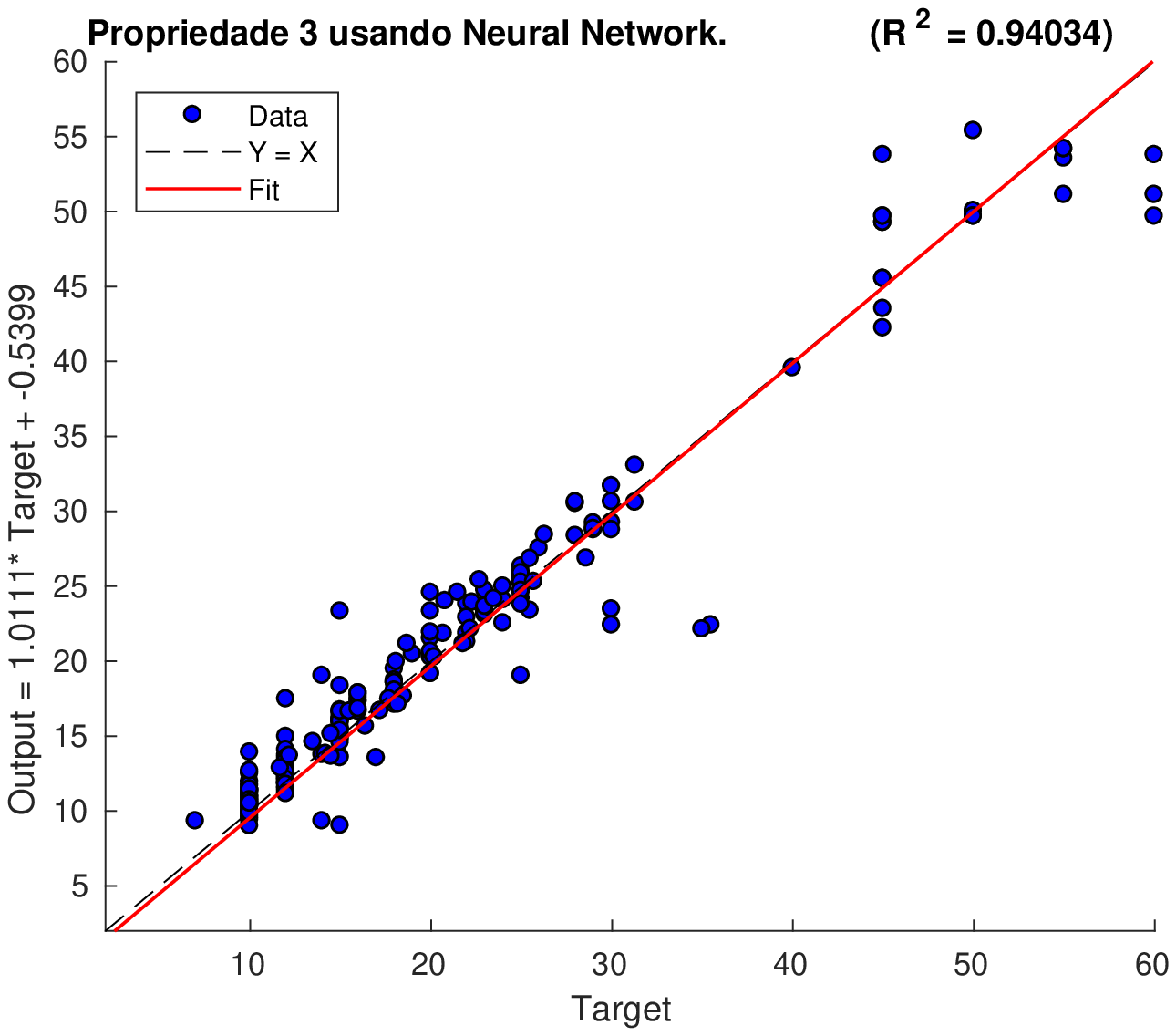} 
\caption{Previsões \textit{vs targets} para a propriedade 3 (\textit{Neural Network}).} \label{FigNN3}
\end{figure}

\begin{figure}
\centering
\includegraphics[scale=0.5]{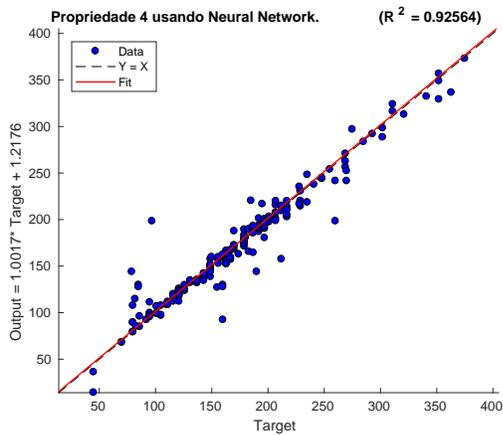}
\caption{Previsões \textit{vs targets} para a propriedade 4 (\textit{Neural Network}).} \label{FigNN4}
\end{figure}

Após obter os $12$ modelos escolhemos o melhor comparando eles via teste de \textit{Friedman-Bonferroni} (\ref{Bonf}). O erro final no conjunto de teste é calculado como sendo a média da sua avaliação em cada um dos $10$ preditores do modelo escolhido. As funções de ativação que melhor trabalharam foram a "Bayesiana" e a de "Levenberg-Marquardt". O coeficiente de determinação meio foi de $0.95$, $0.84$, $0.94$ e $0.92$ para as propriedades $1$, $2$, $3$ e $4$ respetivamente. As Figuras \ref{FigNN1}, \ref{FigNN2}, \ref{FigNN3} e \ref{FigNN4} apresentam o \textit{plot} dos \textit{outputs vs targets}.


\subsection{\textit{SVR}}

Para o treinamento do modelo SVR usamos função do \textit{MATLAB} \textit{fitrsvm}. Treinamos modelos  variando os \textit{kernels} entre \textit{linear, gaussian, rbf e polynomial}. O teste de \textit{Friedman-Bonferrari} (\ref{Bonf}) determinou que os modelos que usaram \textit{Kernel} Gaussiano foram os melhores em cada propriedade. O erro final no conjunto de teste é calculado como sendo a meia da sua avaliação em cada um dos $10$ preditores do modelo escolhido. O coeficiente de determinação meio foi de $0.99$, $0.98$, $0.97$ e $0.98$ para as propriedades $1$, $2$, $3$ e $4$ respetivamente.

\begin{figure}[h!]
\centering
\includegraphics[scale=0.5]{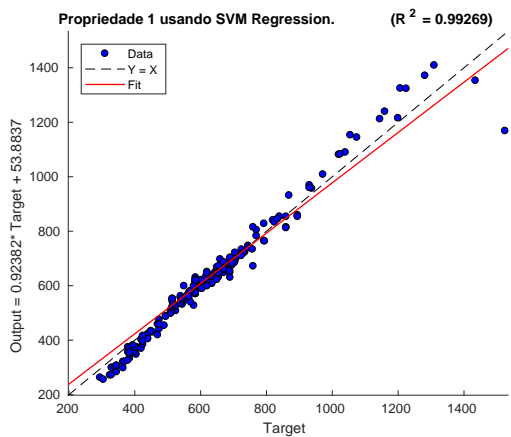}
\caption{Previsões \textit{vs targets} para a propriedade 1 (\textit{SVR}).} \label{FigSVR1}
\end{figure}

\begin{figure}[h!]
\centering
\includegraphics[scale=0.5]{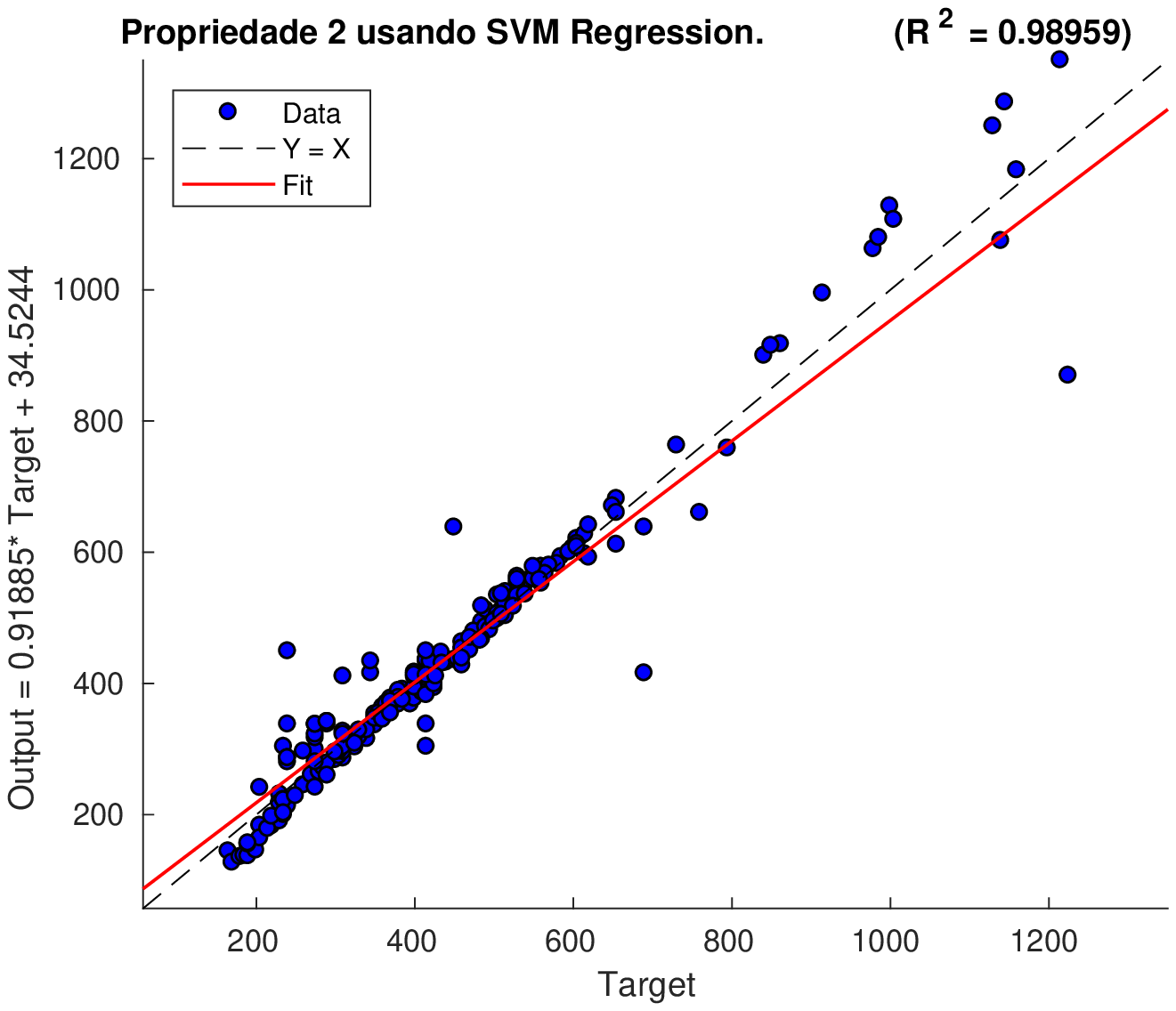}
\caption{Previsões \textit{vs targets} para a propriedade 2 (\textit{SVR}).} \label{FigSVR2}
\end{figure}

\begin{figure}[h!]
\centering
\includegraphics[scale=0.5]{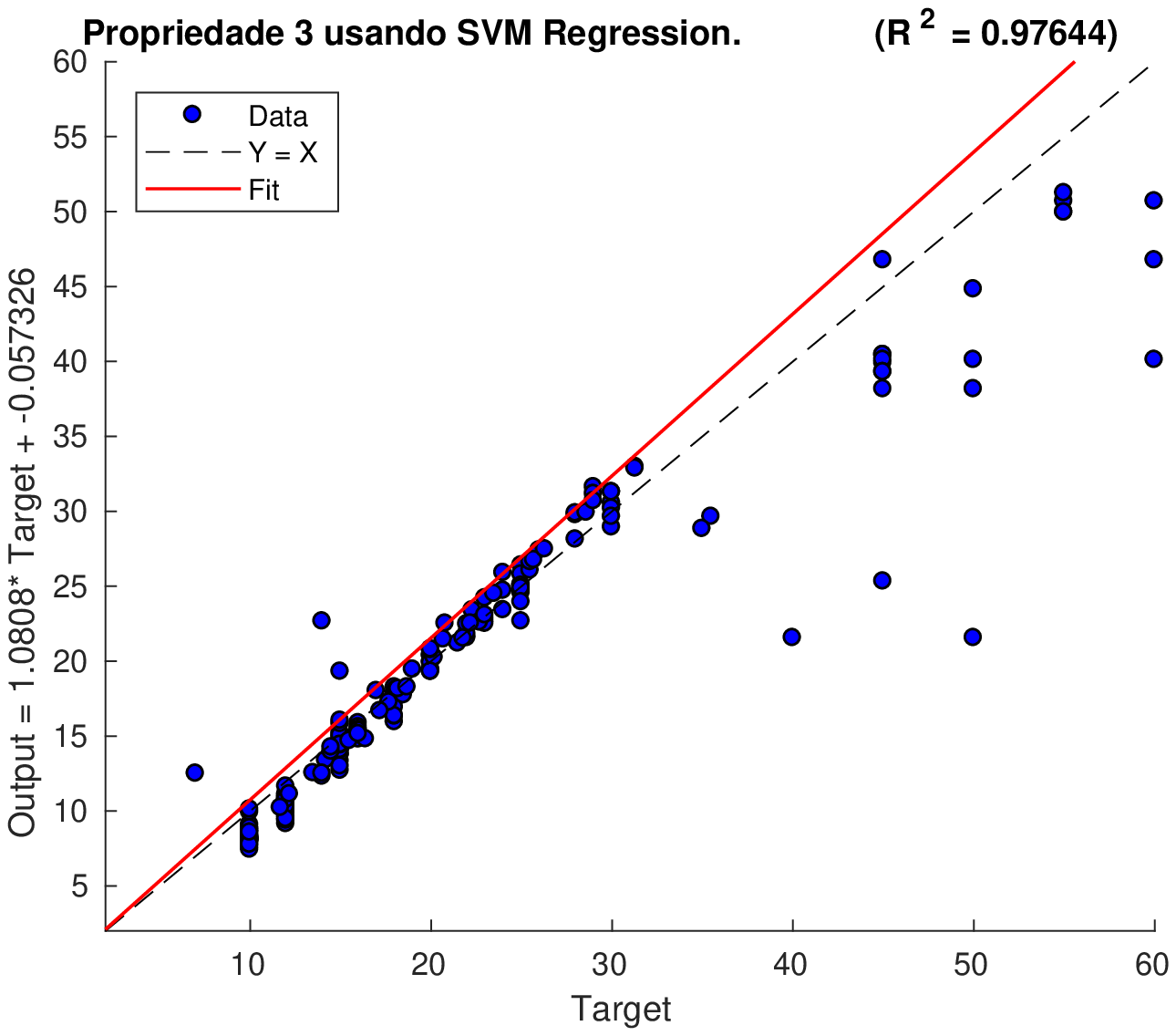}
\caption{Previsões \textit{vs targets} para a propriedade 3 (\textit{SVR}).} \label{FigSVR3}
\end{figure}

\begin{figure}[h!]
\centering
\includegraphics[scale=0.5]{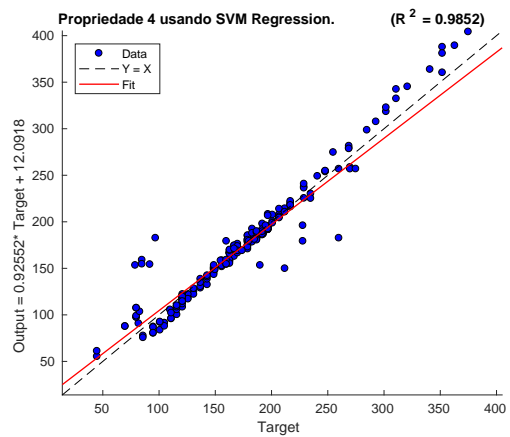}
\caption{Previsões \textit{vs targets} para a propriedade 4 (\textit{SVR}).} \label{FigSVR4}
\end{figure}

 As Figuras \ref{FigSVR1}, \ref{FigSVR2}, \ref{FigSVR3} e \ref{FigSVR4} apresentam o \textit{plot} dos \textit{outputs vs targets}.
 
\newpage
 \
\newpage

\subsection{\textit{Decision Tree}}

\begin{figure}[H]
\centering
\includegraphics[scale=0.5]{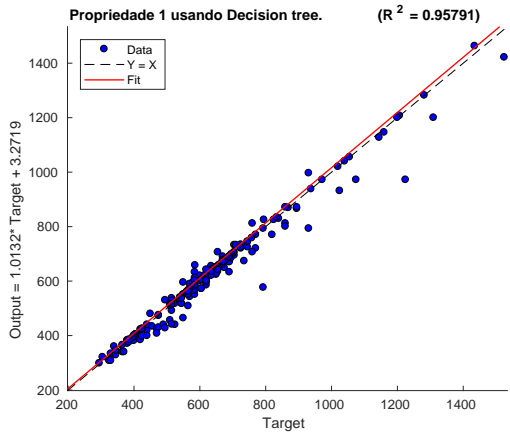}
\caption{Previsões \textit{vs targets} para a propriedade 1 (\textit{Decision Tree}).} \label{FigDec1}
\end{figure}

\begin{figure}[H]
\centering
\includegraphics[scale=0.5]{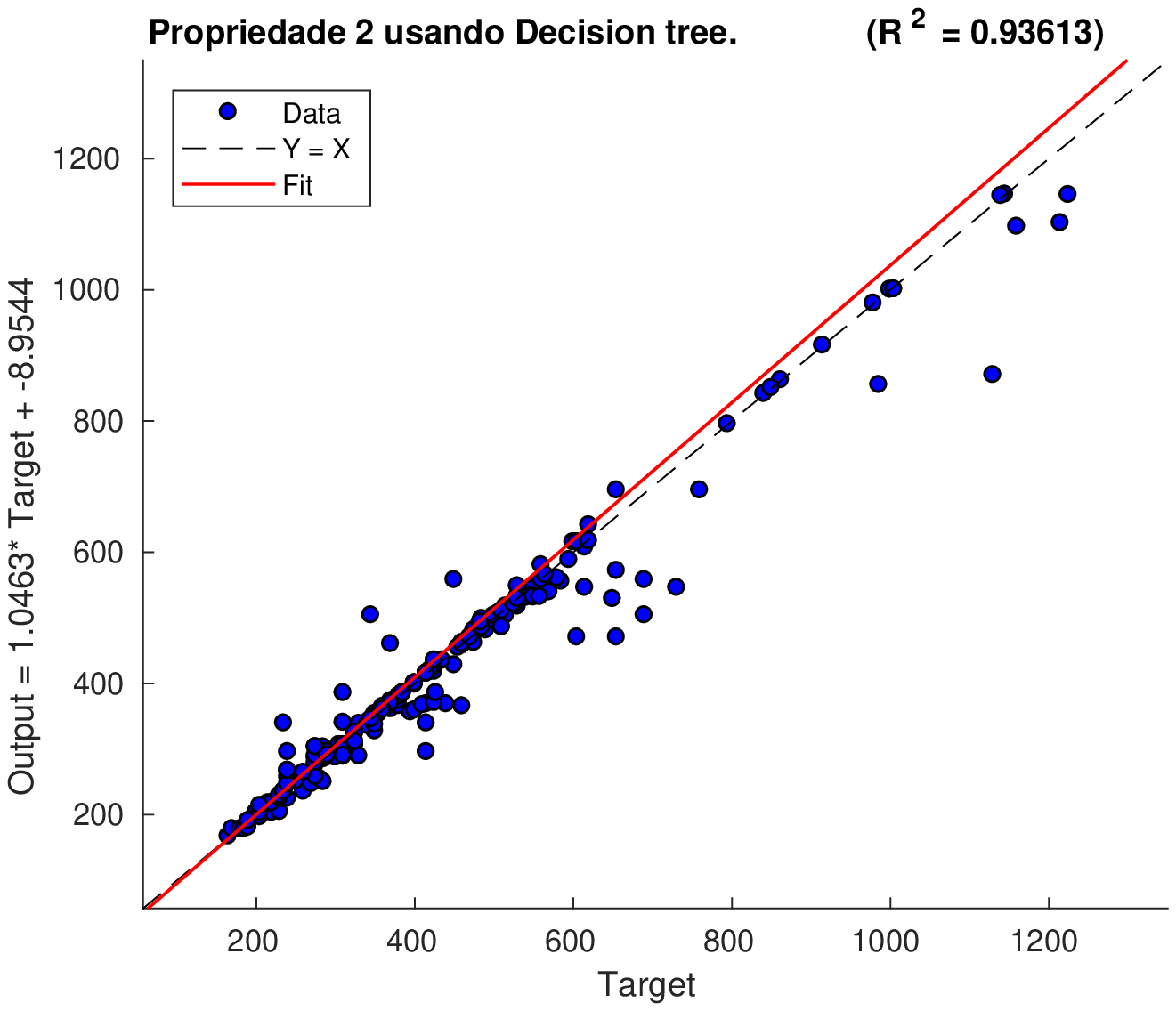}
\caption{Previsões \textit{vs targets} para a propriedade 2 (\textit{Decision Tree}).} \label{FigDec2}
\end{figure}

\begin{figure}[H]
\centering
\includegraphics[scale=0.5]{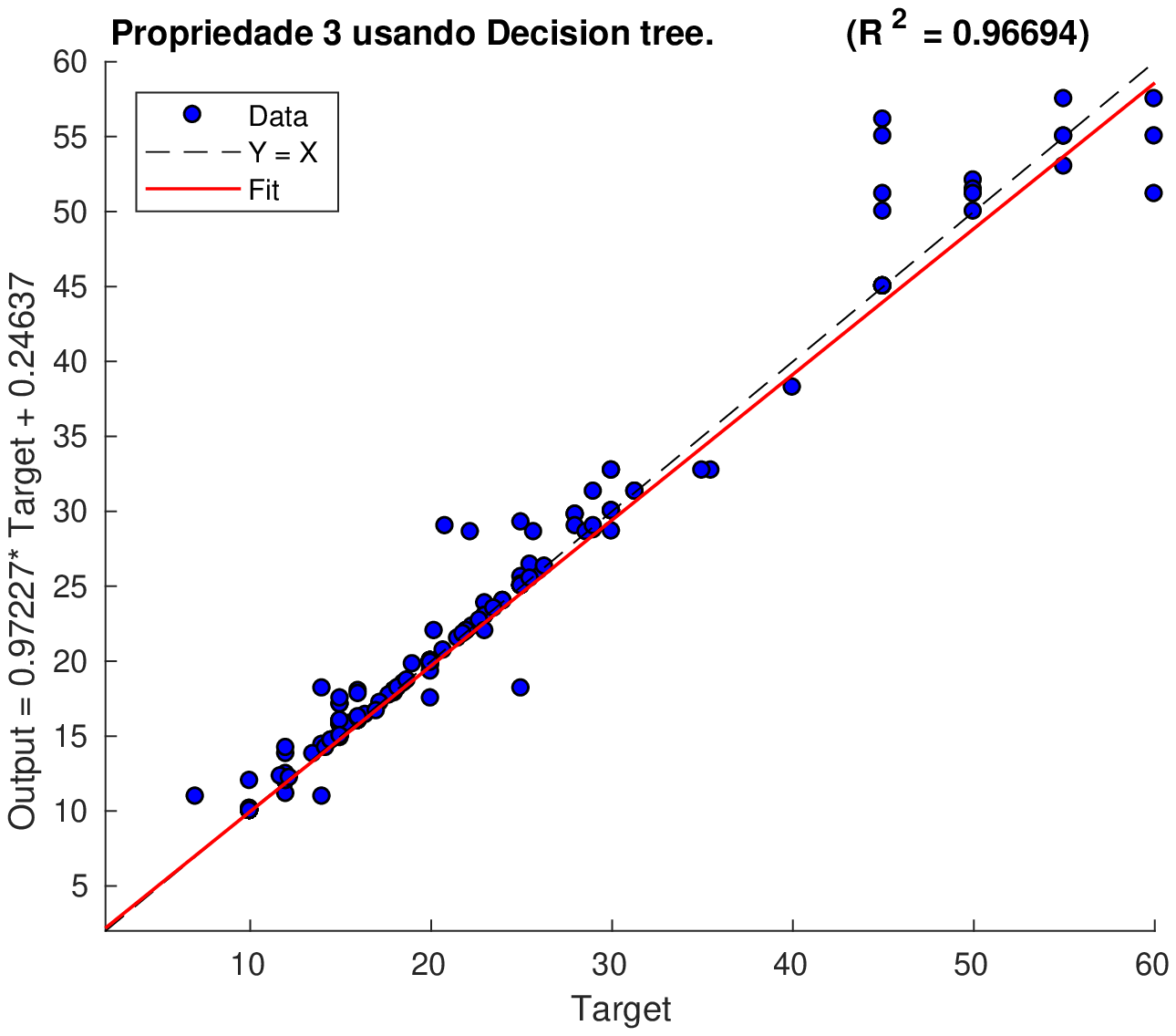}
\caption{Previsões \textit{vs targets} para a propriedade 3 (\textit{Decision Tree}).} \label{FigDec3}
\end{figure}

\begin{figure}[H]
\centering
\includegraphics[scale=0.5]{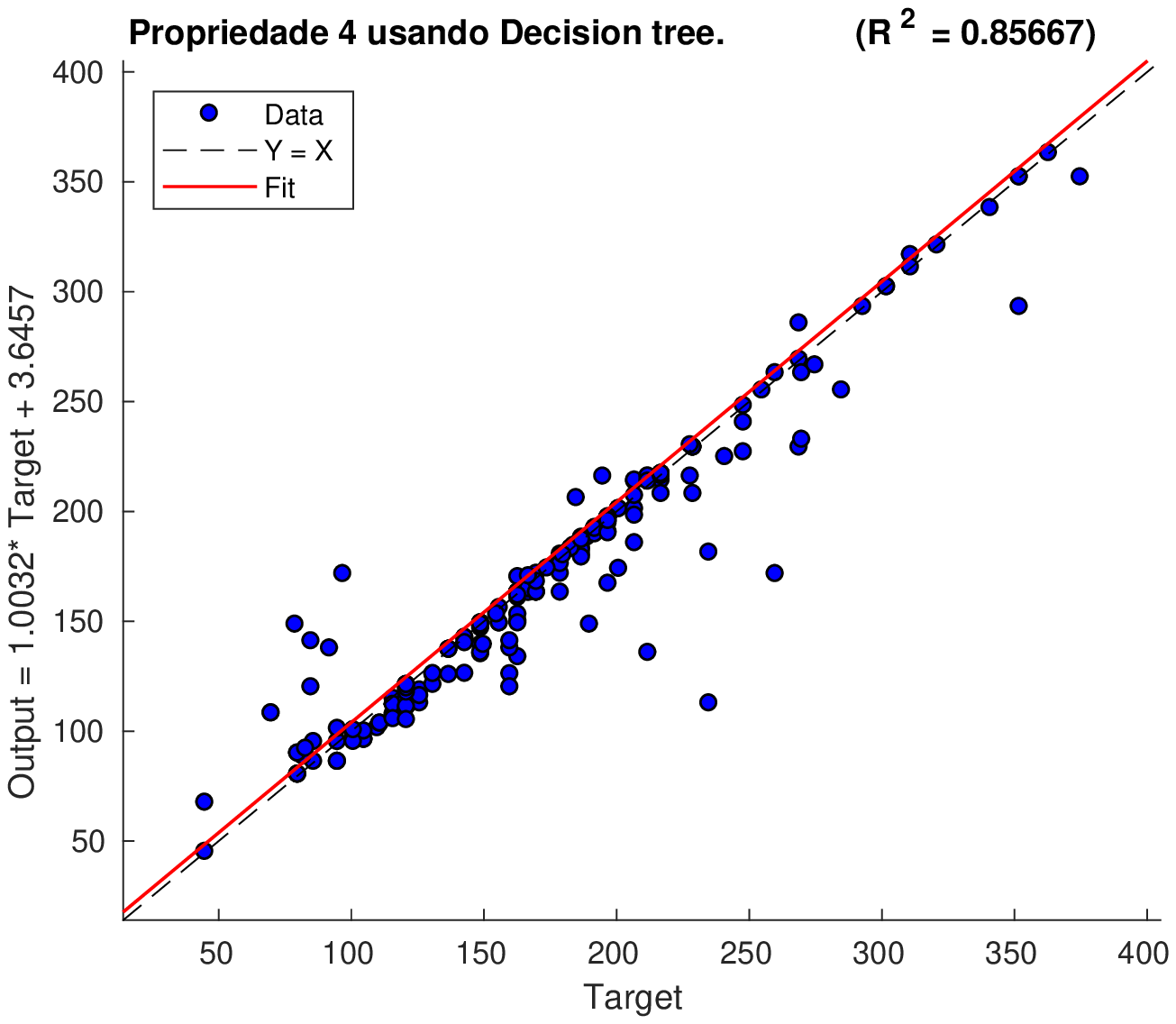}
\caption{Previsões \textit{vs targets} para a propriedade 4 (\textit{Decision Tree}).} \label{FigDec4}
\end{figure}

Para treinar a árvore de regressão usaremos a função de \textit{MATLAB} \textit{fitrtree} e a estratégia de regularização \textit{grow}-\textit{prunning} fazendo crescer a árvore até o final no conjunto de treino e usamos \textit{prunning} até fazer com que o coeficiente de determinação $R^2$ na validação e no treino sejam similares. As Figuras \ref{FigDec1}, \ref{FigDec2}, \ref{FigDec3} e \ref{FigDec4} apresentam o \textit{plot} dos \textit{outputs vs targets}.

\subsection{\textit{Comparando os modelos}} \label{Comp}

  Para cada modelo calculamos o coeficiente de determinação $R^2$ em cada uma das avaliações do conjunto de teste nos $10$ preditores associados.

\begin{table}[!h]
\centering
\scalebox{0.9}{ 
\begin{tabular}{|r|r|r|r|r|r|r|r|r|r|} 
\hline
\rowcolor{cyan}
K-Fold & NN  & SVMR & Dec. Tree & Regre. Lin \\
\hline
\hline
 1 &    0.95533 &    0.99349 &   0.96494 &    0.75425 \\
\hline
\rowcolor{ggray}
 2 &    0.98194 &    0.99383 &   0.96168 &    0.76033 \\
\hline
 3 &    0.95781 &    0.98894 &   0.92343 &    0.74292 \\
\hline
\rowcolor{ggray}
 4 &    0.95485 &    0.99303 &   0.96808 &    0.73856 \\
\hline
 5 &    0.97518 &    0.99283 &   0.96444 &    0.74721 \\
\hline
\rowcolor{ggray}
 6 &    0.87986 &     0.99340 &   0.96082 &    0.76504 \\
\hline
 7 &    0.96407 &    0.99351 &   0.94713 &    0.74609 \\
\hline
\rowcolor{ggray}
 8 &    0.97277 &    0.99286 &   0.96487 &    0.75255 \\
\hline
 9 &    0.91661 &    0.99212 &   0.96443 &     0.75220 \\
\hline
\rowcolor{ggray}
10 &    0.96105 &    0.99284 &   0.95929 &    0.75382 \\
\hline
\hline
Média &    0.95195 &    \textbf{0.99269} &   0.95791 &     0.7513 \\
\hline
\end{tabular}}
\caption{Valores de $R^{2}$ para os modelos e para a propriedade 1 }
\label{Tab1}
\end{table}

\begin{table}[!h]
\centering
\scalebox{0.9}{ 
\begin{tabular}{|r|r|r|r|r|r|r|r|r|r|} 
\hline
\rowcolor{cyan}
K-Fold & NN  & SVMR  & Dec. Tree & Regre. Lin \\
\hline
\hline
 1 &  0.95787 &    0.99066 &    0.94314 &    0.70536 \\
\hline
\rowcolor{ggray}
 2 &  0.57023 &    0.99219 &    0.94121 &    0.72688 \\
\hline
 3 &   0.9103 &    0.97034 &    0.89321 &    0.70381 \\
\hline
\rowcolor{ggray}
 4 &  0.91371 &    0.99347 &    0.94656 &    0.72234 \\
\hline
 5 &  0.66113 &     0.99190 &    0.95071 &    0.71962 \\
\hline
\rowcolor{ggray}
 6 &  0.91158 &    0.99175 &    0.92801 &    0.71392 \\
\hline
 7 &  0.94464 &    0.99315 &    0.92232 &    0.71424 \\
\hline
\rowcolor{ggray}
 8 &  0.75442 &    0.99168 &    0.93227 &     0.72440 \\
\hline
 9 &  0.94637 &    0.98948 &    0.95977 &    0.71998 \\
\hline
\rowcolor{ggray}
10 &  0.84801 &    0.99125 &    0.94415 &    0.72383 \\
\hline
\hline
Média &  0.84183 &  \textbf{0.98959} &    0.93613 &    0.71744 \\
\hline
\end{tabular}}
\caption{Valores de $R^{2}$ para os modelos e para a propriedade 2 }
\label{Tab2}
\end{table}

\begin{table}[!h]
\centering
\scalebox{0.9}{ 
\begin{tabular}{|r|r|r|r|r|r|r|r|r|r|} 
\hline
\rowcolor{cyan}
K-Fold & NN  & SVMR  & Dec. Tree & Regre. Lin \\
\hline
\hline
 1 &   0.93555 &    0.98021 &    0.97349 &    0.88387 \\
\hline
\rowcolor{ggray}
 2 &   0.94766 &    0.98114 &    0.97093 &    0.88236 \\
\hline
 3 &   0.93898 &    0.96998 &    0.96533 &    0.88483 \\
\hline
\rowcolor{ggray}
 4 &   0.93468 &    0.98368 &    0.96103 &    0.88207 \\
\hline
 5 &   0.93687 &    0.96675 &     0.97340 &    0.88473 \\
\hline
\rowcolor{ggray}
 6 &   0.95075 &    0.97357 &    0.96058 &    0.88032 \\
\hline
 7 &    0.93670 &    0.97341 &    0.97251 &    0.88809 \\
\hline
\rowcolor{ggray}
 8 &   0.93807 &      0.97900 &    0.96545 &    0.89123 \\
\hline
 9 &     0.94900 &    0.97561 &       0.96000 &    0.88529 \\
\hline
\rowcolor{ggray}
10 &   0.93512 &    0.98109 &    0.96662 &    0.86381 \\
\hline
\hline
Média &   0.94034 &    \textbf{0.97644} &    0.96694 &    0.88266 \\
\hline
\end{tabular}}
\caption{Valores de $R^{2}$ para os modelos e para a propriedade 3}
\label{Tab3}
\end{table}

\begin{table}[!h]
\centering
\scalebox{0.9}{ 
\begin{tabular}{|r|r|r|r|r|r|r|r|r|r|} 
\hline
\rowcolor{cyan}
K-Fold & NN  & SVMR & Dec. Tree & Regre. Lin \\
\hline
\hline
 1 &    0.93183 &     0.98340 &   0.87735 &    0.76972 \\
\hline
\rowcolor{ggray}
 2 &    0.93847 &    0.99011 &    0.85840 &    0.76531 \\
\hline
 3 &    0.93007 &    0.97645 &   0.83434 &     0.77460  \\
\hline
\rowcolor{ggray}
 4 &    0.90529 &    0.98522 &   0.89335 &    0.77777 \\
\hline
 5 &    0.87144 &    0.98471 &   0.89245 &    0.77831 \\
\hline
\rowcolor{ggray}
 6 &    0.93292 &    0.98386 &   0.88705 &    0.77376 \\
\hline
 7 &    0.93663 &     0.98720 &   0.81248 &    0.77443 \\
\hline
\rowcolor{ggray}
 8 &    0.93723 &    0.98709 &   0.89773 &     0.77560  \\
\hline
 9 &    0.93819 &    0.98286 &     0.71800 &    0.76442 \\
\hline
\rowcolor{ggray}
10 &    0.93431 &    0.99106 &   0.89552 &    0.78311 \\
\hline
\hline
Média &    0.92564 &     \textbf{0.98520} &   0.85667 &     0.77370  \\
\hline
\end{tabular}}
\caption{Valores de $R^{2}$ para os modelos e para a propriedade 4}
\label{Tab4}
\end{table}

   Apenas observando as Tabelas \ref{Tab1}, \ref{Tab2}, \ref{Tab1}  e \ref{Tab4} podemos concluir que o modelo de regressão linear tem uma performance claramente inferior aos outros $3$ em cada uma das propriedades. Por outro lado o modelo SVR ganha praticamente sempre, em todas as propriedades e praticamente em cada cada um dos $10$ \textit{folders}. Não precisamos fazer comparação estatística para concluir que a performance do modelo SVR é melhor do que as dos outros.

\newpage
\section{Conclusões e Trabalhos Futuros}

  Depois das comparações feitas na Seção \ref{Comp} concluímos que o melhor modelo para cada uma das propriedade dentre os $4$ testados foi \textit{SVR} com um coeficiente de determinação médio de $R^2 \approx 0.98 = \frac{1}{4}(0.99 +0.98+ 0.97 +0.98)$) 
  
  O bom do ajuste do modelo aos dados mostra também que as nossas \textit{features} descrevem bem o problema . No caso, as concentrações dos materiais e o tipo de processamento da liga de aço descrevem bem as propriedades de dureza (\textit{hardness}), limite de resistência (\textit{tensile strength}), limite de escoamento (\textit{yield strength}) e ductibilidade (\textit{elongation}).
  
  É muito provável que se tivéssemos um maior número de dados a atuação dos modelos iria melhorar. Também teria sido possível usar outras variantes de \textit{data augmentation} usando algum critério de distribuição dos dados dentro da faixa. Por exemplo poderíamos pensar que a concentração efetiva do material encontra-se no ponto médio da faixa e então gerar os dados aleatoriamente seguindo uma distribuição normal centrada nesse ponto. Outras estratégias de validação cruzada poderiam ter sido usadas, por exemplo \textit{5x2}.
  
  No trabalho tivemos a oportunidade de combinar muitas das técnicas estudadas ao longo do curso para solucionar um problema real de grande importância na indústria. 
  

\end{document}